\documentclass[letterpaper, 10 pt, conference]{ieeeconf}  %

\IEEEoverridecommandlockouts
\usepackage[T1]{fontenc}
\usepackage[utf8]{inputenc}
\usepackage[english]{babel}
\usepackage{csquotes}

\usepackage[pdftex]{graphicx}
\graphicspath{{figures/}}
\usepackage[inkscapearea=page]{svg}
\svgpath{figures/}
\usepackage[font=footnotesize]{caption} %
\usepackage[extractpath=out/svg-extract,extractname=filename]{svg-extract}

\usepackage{amsmath}
\usepackage{siunitx}

\usepackage{hyperref}
\hypersetup{
    colorlinks=true,
    linkcolor=black,
    citecolor=black,
    filecolor=black,
    urlcolor=black,
}
\usepackage[nameinlink, capitalise]{cleveref}

\usepackage[style=ieee,
            doi=false,
            url=false,
            mincitenames=1,
            maxcitenames=1,
            minbibnames=6,
            maxbibnames=6,
            backend=biber]{biblatex}  %
\DefineBibliographyStrings{english}{
    andothers = {et\addabbrvspace al\adddot}
}
\renewbibmacro*{date}{%
  \iffieldundef{year}
    {\bibstring{nodate}}
    {\printdate}}
\usepackage{flushend}

\usepackage[acronym]{glossaries}

\usepackage[colorinlistoftodos]{todonotes}
\usepackage{orcidlink}
\let\labelindent\relax %
\usepackage{enumitem}

\usepackage{customAcronyms}
\usepackage{customMath}
\usepackage{dblfloatfix}
\usepackage{subcaption}

\usetikzlibrary{calc}

\AtBeginBibliography{\small}

\title{%
Human-Aided Trajectory Planning for Automated Vehicles \\
  through Teleoperation and Arbitration Graphs
}

\author{
    Nick Le Large \orcidlink{0009-0006-5191-9043} $^{1*}$,
    David Brecht \orcidlink{0009-0006-6446-2310} $^{2*}$,
    Willi Poh \orcidlink{0009-0007-8217-6438} $^{1}$,\\
    Jan-Hendrik Pauls \orcidlink{0000-0003-2048-392X} $^{1}$,
    Martin Lauer \orcidlink{0000-0003-4414-5722} $^{1}$ and
    Frank Diermeyer \orcidlink{https://orcid.org/0000-0003-1441-5226} $^{2}$
    \thanks{
        $^{*}$ Authors contributed equally to this work
    }
    \thanks{
        $^{1}$ Institute of Measurement and Control Systems, Karlsruhe Institute of Technology (KIT), Karlsruhe, Germany
    }
    \thanks{
        $^{2}$ Chair of Automotive Technology, Technical University of Munich, Munich, Germany
    }
    \thanks{
        $^{3}$ See also \href{https://youtu.be/fVSO-YOeGMk}{\nolinkurl{youtu.be/fVSO-YOeGMk}} for a video demonstration
    }
}

\usepackage{textcomp}
\usepackage[absolute]{textpos}
\usepackage{setspace}

\TPMargin{5pt}

\newcommand*{\presentationfont}{\fontfamily{lmss}\selectfont}
\newcommand{\presentationinformation}[2]{
\TPshowboxesfalse 
\begin{textblock}{0.82}(0.02,0.01)
    \vspace{2mm}
    \noindent{\bfseries{\presentationfont{\footnotesize{#1 \\
                                                        DOI: \href{http://dx.doi.org/#2}{#2}}}}}
\end{textblock}
}

\newcommand{\copyrightstatement}[1]{
    \begin{textblock}{0.825}(0.089,0.94)
        \setstretch{0.65}%
        \noindent
        \begin{minipage}{\linewidth}
            {\fontsize{6.5pt}{8pt}\selectfont
                \parbox{\dimexpr\linewidth-0mm}{%
                    \copyright{} #1 IEEE.
                    Personal use of this material is permitted.
                    Permission from IEEE must be obtained for all other uses, in any current or future media,
                    including reprinting/republishing this material for advertising or promotional purposes,
                    creating new collective works, for resale or redistribution to servers or lists,
                    or reuse of any copyrighted component of this work in other works.
                }
            }
            \vspace{-1mm} %
        \end{minipage}
    \end{textblock}
}

\newcommand*\circled[1]{\tikz[baseline=(char.base)]{
            \node[shape=circle,draw,inner sep=0.5pt] (char) {#1};}}

\begin{document}

\presentationinformation{Presented at the IEEE Intelligent Vehicles Symposium 2025}{10.1109/IV64158.2025.11097737}
\maketitle

\copyrightstatement{2025}

\thispagestyle{empty}
\pagestyle{empty}

\begin{abstract}

Teleoperation enables remote human support of automated vehicles in scenarios where the automation is not able to find an appropriate solution.
Remote assistance concepts, where operators provide discrete inputs to aid specific automation modules like planning, is gaining interest due to its reduced workload on the human remote operator and improved safety.
However, these concepts are challenging to implement and maintain due to their deep integration and interaction with the automated driving system.
In this paper, we propose a solution to facilitate the implementation of remote assistance concepts that intervene on planning level and extend the operational design domain of the vehicle at runtime.
Using arbitration graphs, a modular decision-making framework, we integrate remote assistance into an existing automated driving system without modifying the original software components.
Our simulative implementation demonstrates this approach in two use cases, allowing operators to adjust planner constraints and enable trajectory generation beyond nominal operational design domains.

\end{abstract}
\section{Introduction}
\label{introduction}

\begin{figure}[!t] 
    {\includegraphics[width=1\linewidth]{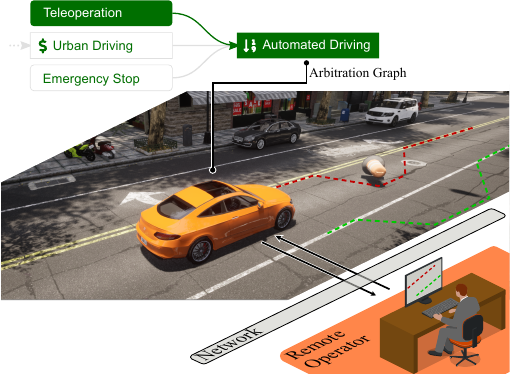}}
    \caption{
        An automated vehicle stops in front of an obstacle blocking the lane, unable to proceed without human intervention.
        In the presented approach, the vehicle automatically senses its need for teleoperation support through the usage of an arbitration graph for decision-making.
        A human remote operator provides guidance by modifying the planning constraints (red and green lines), enabling the vehicle to pass the obstacle.
        This scenario and results are presented in-depth in \cref{sec:results}.
    }
    \label{fig:motivation}
\end{figure}

Level 4 \glspl{AV} have been deployed progressively in commercial and research settings in recent years.
However, \glspl{AV} still encounter situations that cannot be solved by automation alone, leading to disengagements, i.e. deactivation of the \gls{ADS} that prevent mission completion.
An exemplary disengagement scenario is depicted in \cref{fig:motivation} where a object is blocking the \gls{AV}'s lane, forcing it to stop due to the inability to proceed without human support to assess and resolve the scenario.
To address these situations, \gls{AV} companies use teleoperation technology, enabling human \glspl{RO} to assist \glspl{AV} remotely via mobile networks~\cite{Waymo2024,NytZooxArticle2024,NytCruiseArticle2023}.
For example, Cruise required approximately 1.5 workers per vehicle, including remote assistance personnel, intervening every 2.5 miles~\cite{NytCruiseArticle2023}. This highlights the significant role of teleoperation in scaling \gls{AV} fleets and the need for further research in this field.

\begin{figure*}[t]
    \centering
    \includegraphics[width=0.8\linewidth]{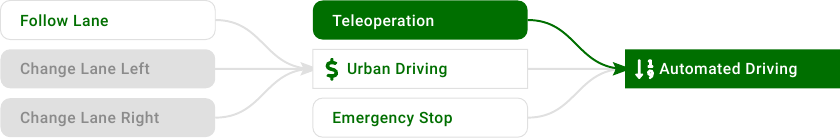}
    \caption{
        The arbitration graph used in this paper during active teleoperation.
        It is a simplified version of the arbitration graph introduced in~\cite{orzechowskiVerhaltensentscheidungFuerAutomatisierte2023} extended by the \behavior{Teleoperation} behavior component.
        Highlighted in green are currently active nodes, grayed out nodes are currently not applicable.
    }
    \label{fig:arbitration-graph-active-teleoperation}
\end{figure*}

Recent approaches to teleoperation focus on remote assistance, where \glspl{RO} support specific \gls{ADS} modules such as perception or planning through high-level instructions instead of controlling the vehicle directly via steering wheel and pedals.
This reduces \gls{RO} workload through event-based rather than continuous actions and increases safety~\cite{Brecht2024EvaluationOfConcepts}.
However, these methods are deeply integrated into specific \gls{ADS} architectures, limiting flexibility and extendability.

To overcome these challenges, we propose a solution that seamlessly integrates human decision-making on planning level using arbitration graphs~\cite{orzechowskiDecisionMakingAutomatedVehicles2020}.
We implement teleoperation as an additional behavior component in the \gls{ADS} stack that enables interaction with a planner without modifying existing planning components.
As shown in \cref{fig:motivation}, when the \gls{AV} encounters unsolvable scenarios, the teleoperation component engages.
The \gls{RO} can now modify planning constraints to allow the \gls{AV} to pass the obstacle under human supervision after which the \gls{RO} hands back control to the \gls{AV}.

\subsection*{Contributions}
\begin{description}[align=left]
    \item[Teleoperation Concept] We extend the \gls{ODD} of an \gls{AV} at runtime using human input on the planning level.
    \item[Seamless Integration] We utilize arbitration graphs to ensure compatibility with an existing \gls{ADS} stack.
    \item[HMI Concept] We enable \glspl{RO} to interact with the \gls{AV}'s environment model and planner constraints.
    \item[Proof-of-Concept] We demonstrate two scenarios solvable only with \gls{RO} assistance in simulation.$^{3}$
\end{description}
\section{Related Work}
\label{sec:related_work}

In an \gls{ADS}, seamless integration of decision-making and trajectory planning layers is essential.
These two components serve distinct yet interconnected roles.
The decision-making layer determines the strategic behavior of the system, selecting maneuvers such as lane changes or speed adjustments based on the current situation.
In contrast, the trajectory planning layer realizes these decisions by generating specific trajectories that satisfy both the behavioral goals and the physical constraints of the system.
If no reasonable trajectory can be found by the \gls{ADS} in a disengagement scenario, a human \gls{RO} can support on planning level which will also be introduced in this chapter.

\subsection{Decision-Making}
\label{sec:decision_making}

Historically, \textbf{\glsentrylongpl{FSM}} have been a popular choice for decision-making due to their simplicity and ease of implementation~\cite{wagnerModelingSoftwareFinite2006,hopcroftIntroductionAutomataTheory2007}.
However, their scalability is limited as the complexity of the system increases, leading to a massive growth in states and transitions.

In contrast, \textbf{\glsentrylongpl{BT}} provide a scalable and transparent structure for behavior generation~\cite{colledanchiseBehaviorTreesRobotics2018}.
They allow for modular design and facilitate the addition of new behaviors without extensive reconfiguration.
Despite these advantages, \glsentrylong{BT} lack inherent safety guarantees as their safety and robustness is heavily dependent on the arrangement of the nodes.
Additionally, the return type of a behavior node is a simple binary or ternary value, which limits the flexibility of the system.

\textbf{Arbitration graphs} are a hierarchical and modular decision-making framework designed for autonomous systems originally developed for robot soccer~\cite{lauerCognitiveConceptsAutonomous2010}.
They have since been successfully applied to autonomous driving~\cite{orzechowskiDecisionMakingAutomatedVehicles2020, orzechowskiVerhaltensentscheidungFuerAutomatisierte2023}.
By combining simple atomic behavior components, they generate complex behavior in a bottom-up approach.
This enables incremental integration of behavior components without modifying existing ones.
\cref{fig:arbitration-graph-active-teleoperation} shows a visualization of an arbitration graph.

Behavior components are the leaf nodes of the arbitration graph and compute a command given the current situation.
Their \glsentrylong{IC} specifies whether a behavior component is applicable in the current situation provided by sensor data or an interpreted environment model.
An activated behavior component can signalize that its execution can be continued using the \glsentrylong{CC}.
The internal nodes, called arbitrators, decide between applicable options depending on the arbitrator's decision policy.
Arbitrators can be nested, allowing for a hierarchical decision-making structure.
The icon in front of the arbitrator's name in \cref{fig:arbitration-graph-active-teleoperation} indicates the decision policy: \arbitrator{Automated Driving} is a priority arbitrator whereas \arbitrator{Urban Driving} makes a decision based on the cost of its options.
Arbitrators and behavior components share a generic interface which supports the integration of complex behavior commands such as trajectories.

As a meta-framework, arbitration graphs offer great flexibility by allowing the integration of diverse underlying planning algorithms into a single decision-making structure.
The modular structure of arbitration graphs allows for great maintainability and extensibility while ensuring a high level of transparency and interpretability.
Safety can be reinforced through a tightly integrated verification step, which ensures that only valid and safe commands are executed~\cite{orzechowskiVerhaltensentscheidungFuerAutomatisierte2023, spiekerBetterSafeSorry2024}.
This combination of flexibility, safety, and robustness makes arbitration graphs ideal for complex and dynamic applications such as autonomous driving and robotics.

\begin{figure*}[hbp]
    \centering 
    \begin{subfigure}{0.33\textwidth}
        \includegraphics[width=\linewidth, trim={4.5cm 3.5cm 4.5cm 3.5cm}, clip]{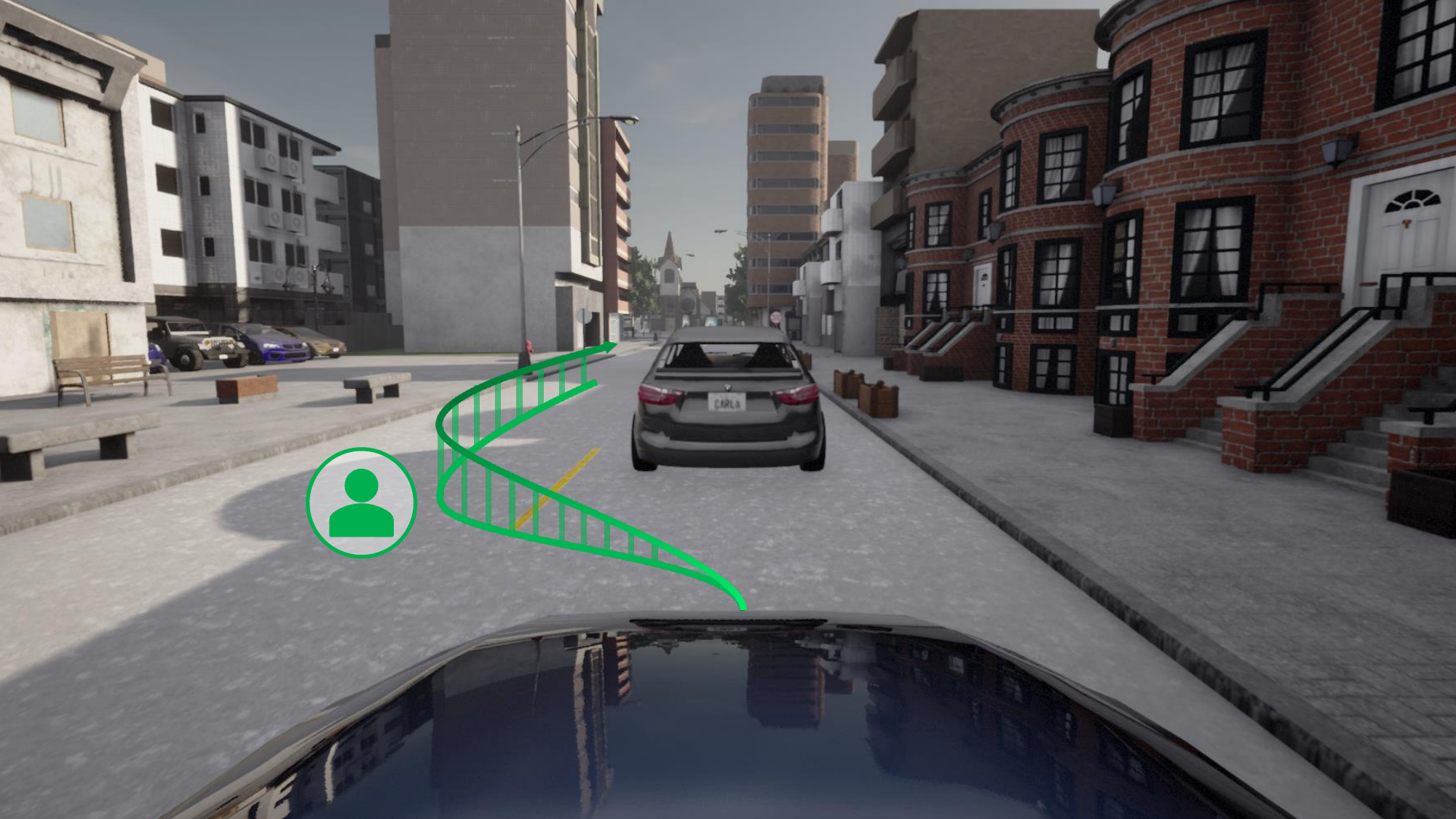}
        \\[-0.3 cm]
        \caption{Trajectory Guidance}
        \label{fig:tg}
    \end{subfigure}%
    \hfill
    \begin{subfigure}{0.33\textwidth}
        \includegraphics[width=\linewidth, trim={4.5cm 3.5cm 4.5cm 3.5cm}, clip]{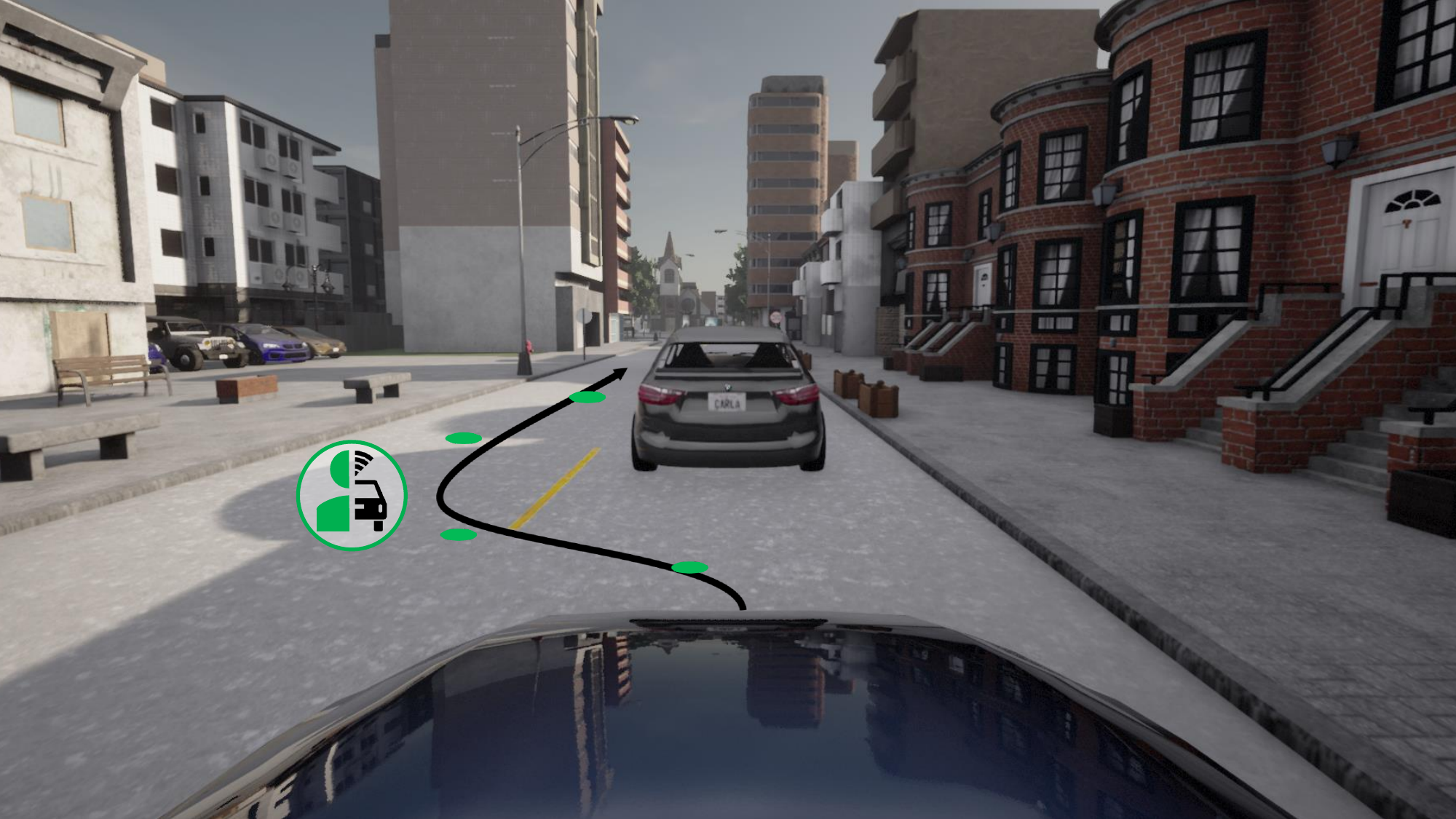}
        \\[-0.3 cm]
        \caption{Waypoint Guidance}
        \label{fig:wg}
    \end{subfigure}%
    \hfill
    \begin{subfigure}{0.33\textwidth}
        \includegraphics[width=\linewidth, trim={4.5cm 3.5cm 4.5cm 3.5cm}, clip]{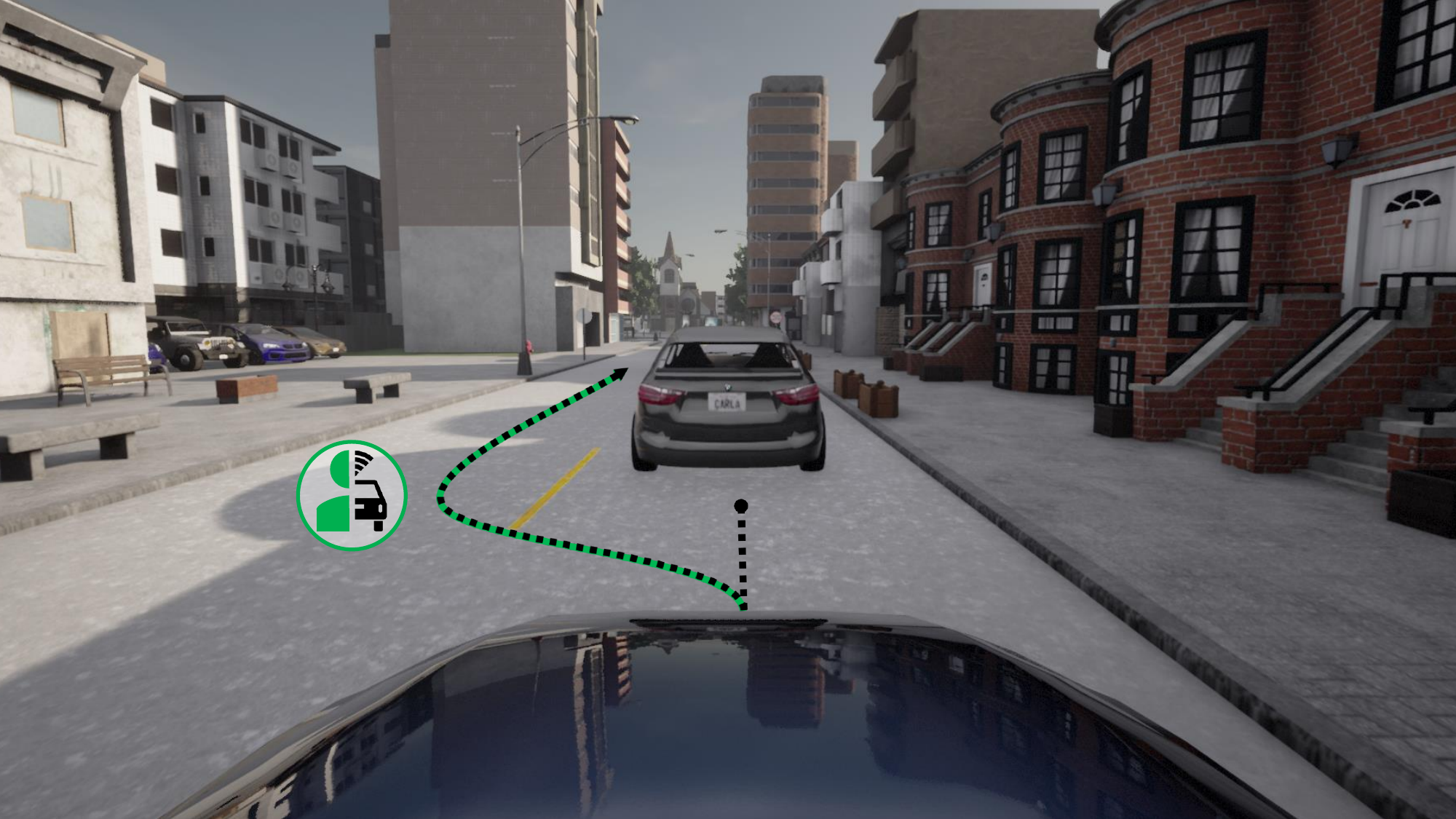}
        \\[-0.3 cm]
        \caption{Collaborative Planning}
        \label{fig:cp}
    \end{subfigure}
    \caption{Overview of trajectory based teleoperation concepts. In trajectory guidance, the \gls{RO} defines all aspects (i.e. curvature and velocity) of the trajectory the \gls{ADS} shall execute. In waypoint guidance, the \gls{RO} inputs waypoints which a planner on the vehicle side takes as input to plan a modified trajectory. In collaborative planning, the \gls{RO} and \gls{ADS} negotiate a trajectory. Figures are taken from~\cite{Brecht2024EvaluationOfConcepts}}
    \label{fig:test}
\end{figure*}

\subsection{Trajectory Planning}

Arbitration graphs enable the integration of multiple trajectory planning algorithms into one \gls{ADS}.
Trajectory planning algorithms are diverse, ranging from rule-based approaches to learned methods.
We refer to \textcite{schwartingPlanningDecisionMakingAutonomous2018} for a detailed overview of trajectory planning methods.

For the \behavior{Teleoperation} behavior component presented in this paper, we use the \gls{MPCC} algorithm as introduced by~\textcite{paulsRealtimeCooperativeMotion2022} which is suitable due to its real-time capability and interpretable constraints.
We are using this algorithm since it is already used by other behavior components of the given system, significantly reducing additional implementation effort.
In the following, we provide a brief overview over the key concepts relevant to this paper.
Refer to the original publication for a more in-depth explanation.

\gls{MPCC} optimizes vehicle motion by maximizing progress along a reference path while minimizing lateral error.
The planning problem for the prediction horizon $N$ is formulated as

\begin{equation}
\min_{z_{1:N+1}} \sum_{k=1}^{N+1} J(\boldsymbol{z}_k),
\end{equation}

subject to system dynamics $\boldsymbol{x}_{k+1} = \mathbf{f}(\boldsymbol{x}_k,\boldsymbol{u}_k)$ for $k~\in~[1,~N]$
and inequality constraints $\boldsymbol{g}(\boldsymbol{z}_k) \leq 0$ for $k \in [1, N+1]$.
The initial state is set to $\boldsymbol{x}_1 = \boldsymbol{x}_\text{init}$.
The input and state variables of stage $k$, $\boldsymbol{u}_k$ and $\boldsymbol{x}_k$, and can be combined into a single vector of stage variables $\boldsymbol{z}_k = [\boldsymbol{x}_k; \boldsymbol{u}_k]$.

A key aspect of MPCC is the introduction of the progress variable $\theta_k$, which represents the arc length of the vehicle along the reference path.
It is part of the state vector $\boldsymbol{x}_k$, its derivative $\dot{\theta}_k$ is included in the control vector $\boldsymbol{u}_k$.

\begin{figure}[!h]
    \centering
    \includegraphics[width=\columnwidth]{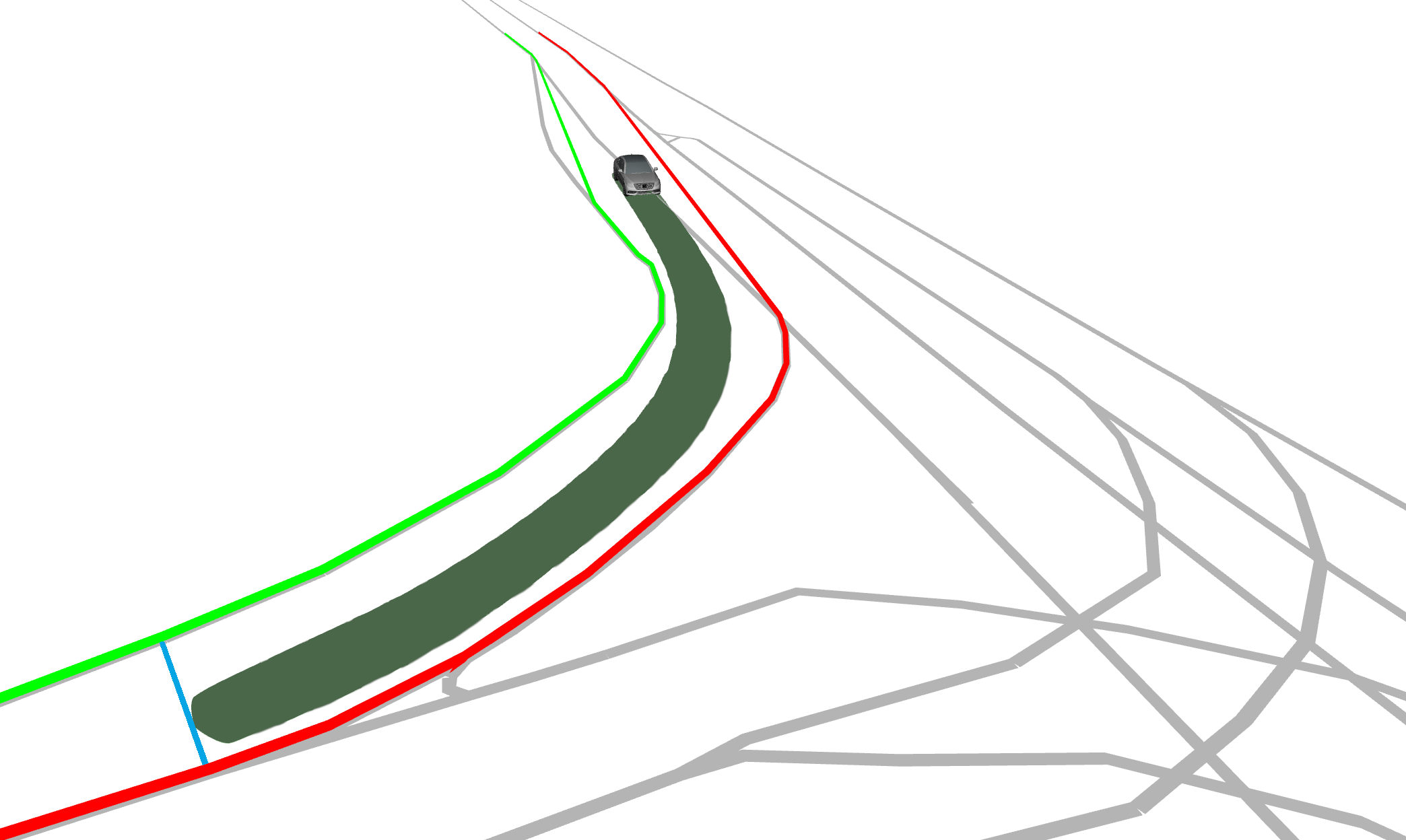}
    \caption{
        The simulation environment used to evaluate the presented concept.
        It shows the ego vehicle during a lane change.
        Grey lines represent the lane topology.
        The lateral constraints are visualized as a red ($E_c^\text{left}$) and a green ($E_c^\text{right}$) line.
        The blue line represents the longitudinal constraint $\theta_\text{stop}$.
        The resulting trajectory is shown in dark green.
    }
    \label{fig:simulation}
\end{figure}

The inequality constraints include the \textbf{lateral constraints}

\begin{equation}
    \boldsymbol{g}_{\text{lat},k} = 
    \begin{bmatrix}
    g_{\text{lat},k}^\text{left} \\
    g_{\text{lat},k}^\text{right}
    \end{bmatrix}
    =
    \begin{bmatrix}
    - E_c^\text{left}(\theta_k) + R_c + \hat{E}_c(\boldsymbol{x}_k) \\
    E_c^\text{right}(\theta_k) + R_c - \hat{E}_c(\boldsymbol{x}_k)
    \end{bmatrix}
\end{equation}

which ensure that the vehicle remains within the drivable corridor.
The boundaries $E_c^\text{left}(\theta_k)$ and $E_c^\text{right}(\theta_k)$ define the feasible lateral region, cf.~\cref{fig:simulation}.
The lateral contouring error $\hat{E}_c(\boldsymbol{x}_k)$ approximates the deviation of the vehicle’s center from the reference path.
The vehicle geometry is approximated using circular primitives with radius $R_c$.

Movement along the path can be limited by a \textbf{longitudinal constraint}

\begin{equation}
    g_{\text{lon},k} = \theta_k - \theta_\text{stop}
\end{equation}

which prevents the vehicle from exceeding a predefined progress limit $\theta_\text{stop}$, e.g., at a stop line.

\begin{figure*}[!b]
    \centering
    \includegraphics[width=1\linewidth]{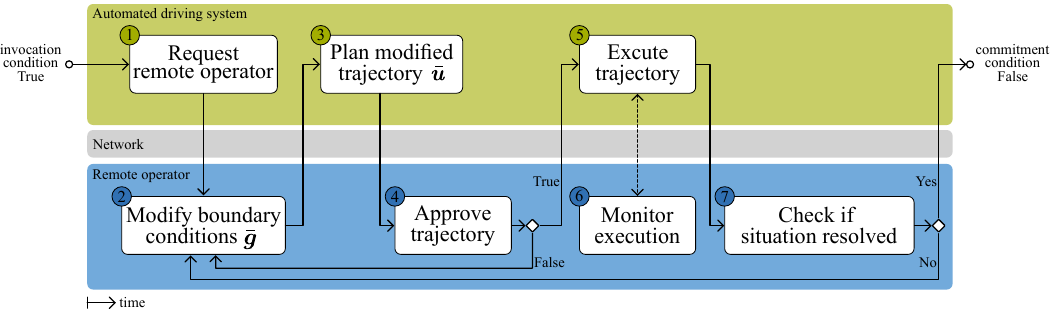}
    \caption{Overview of interaction process between remote operator and automated driving system. The process starts with the \glsentrylong{IC} becoming True if a disengagement is present. The remote operator is requested and supports the automation by providing a set of modified boundary conditions to the planner that plans a modified trajectory the remote operator approves. After trajectory execution, the remote operator checks if the situation was resolved, ending the teleoperation process by manually setting the \glsentrylong{CC} to False.}   
    \label{fig:teleoperation_process}
\end{figure*}

\subsection{Teleoperation on Trajectory Level}
\label{sec:teleoperation-sota}
If no feasible solution to a specific scenario can be found by the \gls{ADS}, a \gls{RO} can support the system remotely.
\textcite{Maj2022bSurveyTeleoperationConcepts} proposed a classification of teleoperation concepts into different classes according to how the \gls{RO} interacts with the \gls{AV}.  
Three teleoperation concepts that assist the \gls{ADS} on trajectory level can be distinguished and are introduced in the following.

The first concept to remotely interact on trajectory level is \textbf{trajectory guidance} where the \gls{RO} defines all parameters of the trajectory, i.e. curvature and velocity, the vehicle shall follow, cf. \cref{fig:tg}.
This can be realized using a steering wheel and pedals~\cite{Gna2012TrajectoryBasedSharedAutonomy, Hof2022bSafeCorridor} or by inputting a set of waypoints with input devices such as mouse and keyboard~\cite{Wolf2024TrajectoryGuidance, Majstorovic2024trajectoryGuidance}. 
In \textbf{waypoint guidance}, the \gls{RO} defines a set of waypoints that the planner of the \gls{ADS} takes as an input to plan a new trajectory that allows to solve the disengagement scenario, cf. \cref{fig:wg}~\cite{5gcroco2021, Gontscharow2024WaypointGuidance}.
In \textbf{collaborative planning}, cf. \cref{fig:cp}, the \gls{RO} provides high-level input, such as modifying the drivable space~\cite{Schitz2021InteractivePathPlanning} or adjusting specific ODD parameters~\cite{Maj2023CollaborativePathPlanning}. Based on this input, the vehicle's planner generates a new trajectory to resolve the disengagement.

Collaborative planning offers various advantages over trajectory guidance and waypoint guidance, including lower mental demand, reduced risk from erroneous human input and robustness against network instability~\cite{Brecht2024EvaluationOfConcepts}.
These advantages mainly stem from the \gls{RO} interacting directly with the planning module. 

However, the tight coupling of collaborative planning with the \gls{ADS} software modules increases system complexity and limits applicability across different \gls{ADS} architectures. 
To address this issue and facilitate easy implementation of teleoperation solutions into the \gls{ADS} software stack, we use arbitration graphs.
While collaborative planning is not the only approach suitable for integrating teleoperation into the arbitration graph framework, we adopt it due to its benefits outlined above. 
Additionally, this approach allows to utilize the existing planner, further simplifying integration while preserving the advantages of collaborative planning.
\newpage
\section{Methodology}
\label{sec:methodology}

\subsection{Concept Architecture}
\label{sec:concept_architecture}
To integrate teleoperation and human decision-making into the \gls{ADS}, we use arbitration graphs as introduced in \cref{sec:decision_making}.
We choose this approach since it offers a high level of flexibility and modularity over other decision-making frameworks.
By using arbitration graphs, there is no need to modify the existing decision-making stack when adding new behavior components.
In this work, we introduce a \behavior{Teleoperation} behavior component to an existing, real-world proven arbitration graph for automated driving as introduced by~\citeauthor{orzechowskiDecisionMakingAutomatedVehicles2020}~\cite{orzechowskiDecisionMakingAutomatedVehicles2020, orzechowskiVerhaltensentscheidungFuerAutomatisierte2023}.
The resulting graph is displayed in \cref{fig:arbitration-graph-active-teleoperation} in the state of active teleoperation.
For simplicity, fallback layers are not shown in the figure.

As can be seen in the figure, the \behavior{Teleoperation} behavior component is integrated as the first option of the root arbitrator.
This ensures that the \behavior{Teleoperation} behavior component is always activated when the \gls{ADS} cannot find a solution no matter what other behavior components are currently active or applicable.
The \glsentrylong{IC} defines under which prerequisites a behavior becomes applicable.
It should be designed carefully for the \behavior{Teleoperation} behavior component to only activate in disengagement scenarios.
Additionally, the \glsentrylong{CC} defines under which circumstances the activated behavior stays applicable even if the \glsentrylong{IC} is no longer met.
For the \behavior{Teleoperation} behavior component, these conditions are defined as follows:

The \textbf{\glsentrylong{IC}} is fulfilled if a disengagement scenario is present, i.e. the \gls{ADS} is not able to find a valid solution, deactivates and needs human intervention through a \gls{RO} to solve the scenario.
For this work, we assume this to be the case when the vehicle is not making progress towards its goal and is therefore failing to complete its mission.
This is the case when the vehicle is at a standstill for a certain amount of time,
i.e. $v_\text{ego} = \SI{0}{\metre\per\second}$ and $\Delta t_\text{standstill} \geq \Delta t_\text{invocation}$
where $\Delta t_\text{standstill}$ is the time the vehicle has been at a standstill and $\Delta t_\text{invocation}$ is a threshold parameter.
Further, situation-aware checks are possible and sensible but out of scope for this work.
For example, $\Delta t_\text{invocation}$ could be higher near a traffic light or in a traffic jam.

The \textbf{\glsentrylong{CC}} is fulfilled until the \gls{RO} sends a specific request to deactivate the \behavior{Teleoperation} behavior component.
This ensures that the vehicle does not return to fully automated operation without explicit human approval.
\subsection{Teleoperation Process}
\label{sec:teleoperation_process}
The process starts with the \behavior{Teleoperation} behavior becoming applicable via its \glsentrylong{IC} as described above, cf. \cref{fig:teleoperation_process}.

\begin{figure}[!h]
    \centering
    \includegraphics[width=\linewidth]{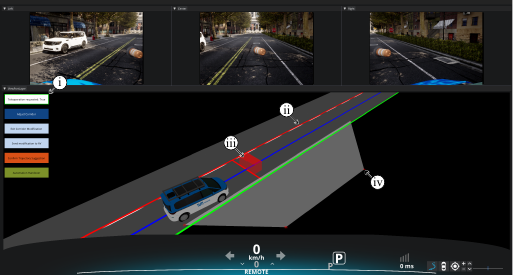}
    \caption{HMI enabling the \gls{RO} to interact with the \gls{ADS} and modify planner boundary conditions. The top section displays video streams of the current scenario, while the bottom section presents an interactive environment model of the vehicle. The buttons on the left provide control options for adaption of the planner's boundary conditions, trajectory approval and relinquishing control to the \gls{ADS}.}
    \label{fig:operator-hmi}
\end{figure}

In the initial step of the process, the \gls{ADS} initiates a request for remote support from the \gls{RO} \circled{1}. 
The \gls{RO}, located in a remote control center, is notified of the teleoperation request, through an \gls{HMI} as illustrated in \cref{fig:operator-hmi} \circled{i}.
The \gls{RO} interacts with the \gls{HMI} that presents live camera streams in the upper section and the \gls{ADS} environment model along with planner constraints $\boldsymbol{g}$ \circled{ii} in the lower section. 
This configuration enables the \gls{RO} to gain understanding of the vehicle’s surroundings and develop situational awareness, including identification of the cause for the teleoperation request. 
In the depicted scenario, a traffic cone being recognizes as a blocking obstacle and obstructing the vehicle's lane can be identified as the source of the disengagement \circled{iii}.

The \gls{HMI} design draws on established research and industry practices in remote driving and assistance, e.g. \cite{Waymo2024, NytZooxArticle2024}, and integrates selected elements like the lower status bar that were implemented and systematically evaluated in \cite{wolf2025hmi}.
The \gls{HMI} components are available as open source software \cite{kerbl2025tumTeleoperation}.

Using the \gls{HMI}, the \gls{RO} modifies the planner's active constraints  $\bar{\boldsymbol{g}}$ which are then sent to the \gls{ADS} over the network \circled{2}.
In the example depicted in \cref{fig:operator-hmi}, the \gls{RO} extends the lateral constraints through a polygon (grey shaded area \circled{iv}). 
Since planning constraints are modified here, we are classifying this approach as collaborative planning based on the descriptions from \cref{sec:teleoperation-sota}.
On the vehicle side, the planner uses the modified constraints to plan a new trajectory $\bar{\boldsymbol{u}}$ \circled{3}.
This trajectory is sent as a proposal to the \gls{RO} for approval \circled{4}.
While waiting for the \gls{RO}'s adjustment and confirmation, the planning module sends a standstill trajectory to the \gls{ADS}'s controller to keep the vehicle at its current position.
If the trajectory is not approved by the \gls{RO} since it is not valid or desirable, they are required to modify the planner constraints again, triggering computation of another trajectory by the \gls{ADS}.
If the approval is given by the \gls{RO}, the \gls{ADS} executes the defined trajectory \circled{5} under continuous monitoring of the \gls{RO} \circled{6}.
Continous monitoring by the \gls{RO} is needed since the vehicle is outside its nominal \gls{ODD} during teleoperation.
Thereby, if the \gls{ADS} state or trajectory is not deemed valid during execution, the \gls{RO} is able to intervene by stopping tracejtory execution, leading to a standstill of the vehicle.
The HMI was developed based on industry standards for remote assistance of automated vehicles as found in compaies like Waymo or Zoox as well as research on \glspl{HMI} for teleoperation \cite{wolf2025hmi}. The \gls{HMI} is available open source \cite{kerbl2025tumTeleoperation}.

After solving the situation that led to the disengagement, other behavior components such as  \behavior{Follow Lane}, cf. \cref{fig:arbitration-graph-active-teleoperation} should provide reasonable trajectories that are displayed in the \gls{HMI}.
If deemed valid by the \gls{RO}, they send a signal to hand over control back to the \gls{ADS} \circled{6}.
This deactivates the \behavior{Teleoperation} behavior component's \glsentrylong{CC}, at which point it is no longer applicable and a different behavior component is selected by the \gls{ADS}.
\section{Results}
\label{sec:results}

As a proof-of-concept, the concept introduced in \cref{sec:methodology} is evaluated in two exemplary scenarios in simulation.
In both scenarios, the \gls{AV} has to stop in front of a detected obstacle on a one lane road.
In \textbf{scenario~A}, the obstacle turns out to be a smoking manhole cover.
In contrast, the obstacle in \textbf{scenario~B} is concrete barrier that has fallen off a cargo truck.
It cannot be passed while staying compliant to traffic rules.

\begin{figure}[!h]
    \centering
    \includegraphics[width=\columnwidth]{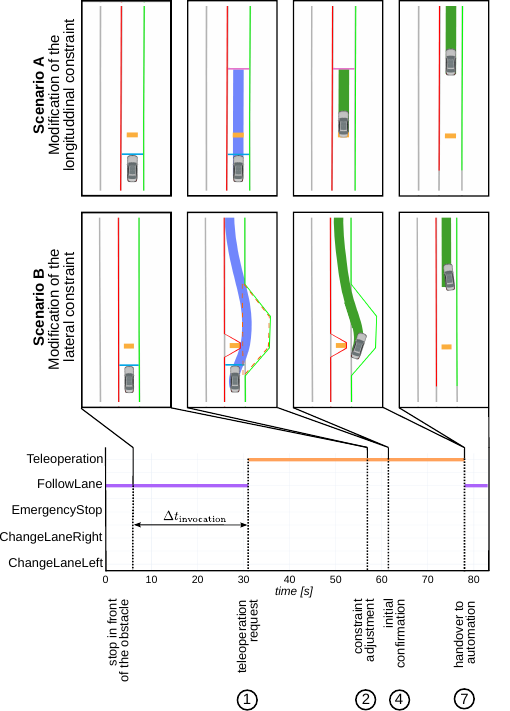}
    \caption{
        Simulated scenarios A and B.
        The timeline at the bottom shows the active behavior component with relevant events being highlighted.
        The screenshots above show key moments during the simulations of scenario~A and B, respectively.
        The dashed polygon in scenario~B represents the modification sent by the \gls{RO}.
        Note that the timeline corresponds to scenario~A but since the events occur analogously, it represents both scenarios.
    }
    \label{fig:simulation-results}
\end{figure}

The simulation seen in \cref{fig:simulation-results} is based on the open source simulation environment CoInCar-Sim~\cite{naumannCoInCarSimOpenSourceSimulation2018}
with realistic lane geometries using our real-world test track in the city of Karlsruhe, Germany.
Both scenarios start with the \gls{AV} executing the \behavior{Follow Lane} behavior and coming to a stop before the detected obstacle.
\cref{fig:simulation-results} gives an overview of the simulation results.

\subsection*{Scenario A: Modification of \textbf{longitudinal} constraint}

\begin{figure}
    {\includegraphics[width=1\linewidth]{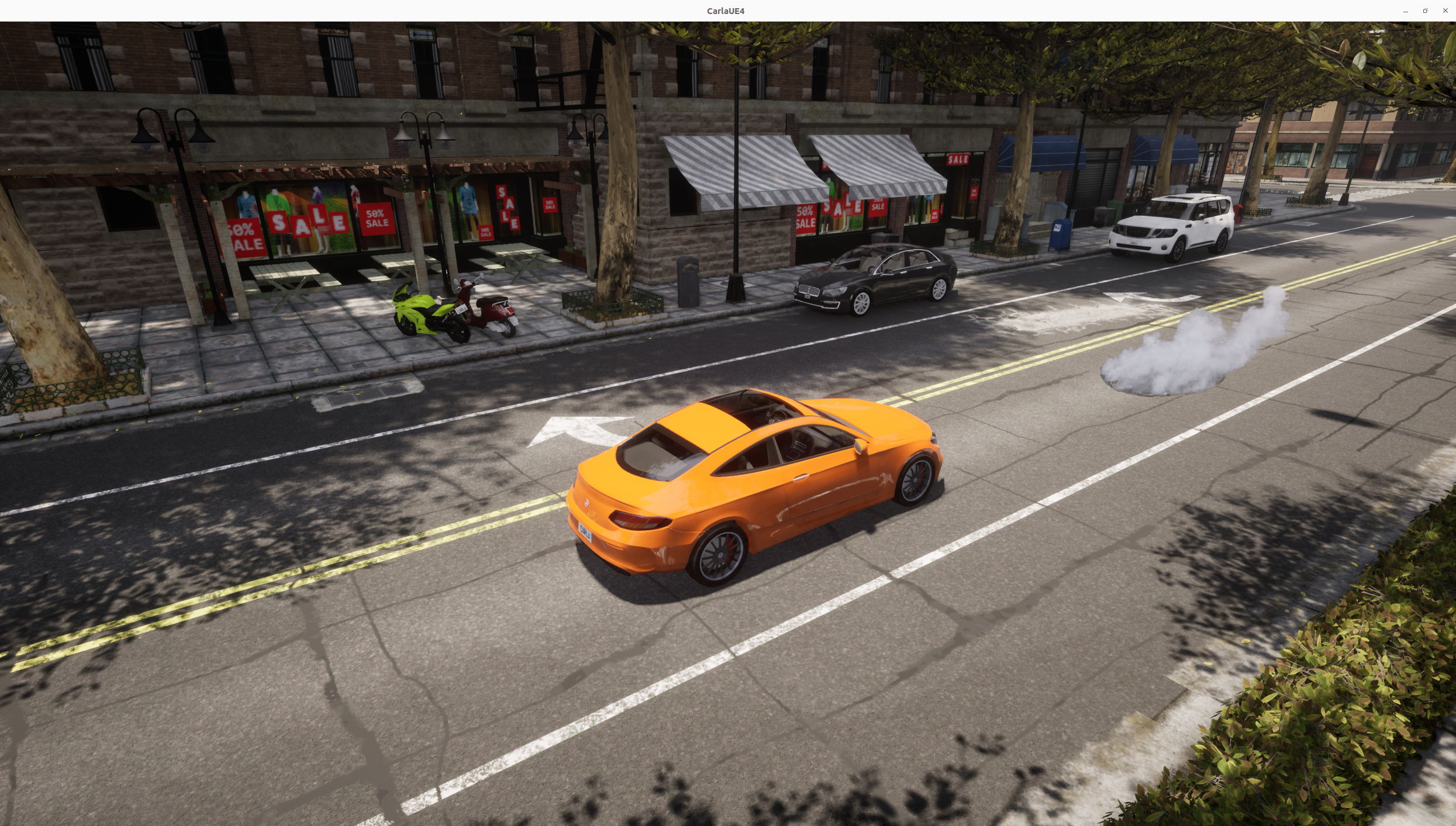}}
    \caption{
        In scenario~A, the \gls{AV} stops in front of a smoking manhole cover.
        Though perceived as an obstacle, the \gls{AV} could pass it safely.
        The \gls{RO} adjusts the longitudinal constraint, allowing the \gls{AV} to continue its journey.
    }
    \label{fig:scenario_a}
\end{figure}

In the first scenario, the perception system of the \gls{AV} detects a smoking manhole cover as an obstacle that blocks the road, cf. \cref{fig:scenario_a}.
At $t=\SI{0}{\second}$, the \behavior{Follow Lane} behavior component is active, stopping the vehicle in front of the obstacle at $t=\SI{6}{\second}$.
After $\Delta t_\text{invocation}=\SI{25}{\second}$ in standstill, the \behavior{Teleoperation} behavior component is applicable.
As the highest priority of the root arbitrator, it is selected at $t=\SI{31}{\second}$ over the \behavior{Follow Lane} behavior component and a \gls{RO} is requested for assistance.

While the \behavior{Teleoperation} behavior component sends a standstill trajectory to the vehicle controller, the \gls{RO} builds situational awareness using the vehicle's environment model and currently active planner constraints $\boldsymbol{g}$.
After determining that the perceived obstacle can be safely ignored, the \gls{RO} sends an adjusted longitudinal constraint $\bar{g}_\text{lon}$ to the \gls{AV} at $t=\SI{57}{\second}$.
This allows planning a trajectory from the vehicle's current position \textit{through} the perceived obstacle to the modified longitudinal constraint $\bar{g}_\text{lon}$ which is then sent to the \gls{RO} as a proposal.
Since the \behavior{Teleoperation} behavior component is committed until the \gls{RO} hands back control to the \gls{AV}, the modified constraint stays active until then.

After the \gls{RO} has confirmed the trajectory at $t=\SI{61}{\second}$, it is sent to the vehicle controller and executed by the \gls{AV}.
During the maneuver, the \gls{RO} carefully monitors the vehicle to ensure that the trajectory is executed as planned.
To avoid issues caused by connection errors, the confirmation signal is sent at a continuous rate.

Once the vehicle has passed the obstacle, the \gls{RO} sends a signal to handover control back to the \gls{ADS} at $t=\SI{78}{\second}$ which deactivates the \glsentrylong{CC} of the teleoperation behavior so that it is no longer applicable.
The next best option to the \arbitrator{Automated Driving} arbitrator is the \arbitrator{Urban Driving} arbitrator which selects its only applicable option, the \behavior{Follow Lane} behavior component.
The vehicle continues its journey in fully automated mode.

\subsection*{Scenario B: Modification of \textbf{lateral} constraint}

Depicted in \cref{fig:motivation} is scenario~B, where a solid obstacle blocks the current lane.
It is properly detected by the \gls{AV} forcing it to stop.
Additionally, the road topology does not allow lane changes due to the solid lane markings on either side of the \gls{AV}'s current lane.

Analogously to scenario~A, the teleoperation behavior component is activated after $\Delta t_\text{invocation}=\SI{25}{\second}$ in standstill.
Again, the vehicle sends a standstill trajectory to the vehicle controller while the \gls{RO} evaluates the situation using the data received from the vehicle.

This time, the \gls{RO} determines that the obstacle cannot be passed without leaving the lane. %
After careful evaluation, the \gls{RO} adjusts the lateral constraints to include the shoulder as drivable area.
For this, the \gls{RO} defines a polygon to the right of the current boundaries that is fused with the existing lateral constraints $g_\text{lat}$ using the area union operation.
The modified lateral constraint $\bar{g}_\text{lat}$ allows the planner to find a trajectory proposal that passes the obstacle using the shoulder.
After confirming the trajectory suggested by the \gls{ADS} based on the modified boundary conditions, the vehicle starts following the trajectory.
Once back in lane, the \gls{RO} hands back the control to the \gls{AV}, which is concluding the teleoperation process and switches to the \behavior{Follow Lane} behavior component.
\section{Conclusion and Future Work}
\label{sec:conclusion}

We presented a concept for integrating teleoperation elegantly into \glspl{AV} using arbitration graphs, seamlessly enabling collaborative human intervention at the planning level.
By allowing remote operators to modify planner constraints, the \gls{AV}'s \gls{ODD} can be extended at runtime.
This enables handling disengagement scenarios without extensive architectural changes.
We demonstrated the effectiveness of the concept in simulation by resolving scenarios such as false-positive obstacles and blocked lanes using external human input.

In this work we focused on the core concept, demonstrating it through two representative scenarios in simulation.
The use of arbitration graphs for teleoperation is easily extensible to a wide range of scenarios and requirements.
As we have shown, the integration of a teleoperation behavior does not require modifications of the existing planning stack. The transitions into the teleoperation behavior and back are fully supported by the existing arbitrators. 
This paves the road for even more complex ways of collaborative teleoperation. For example, the \gls{AV} could be rerouted during emergencies or could adapt to lane topology changes caused by construction work. All this is possible without adjustments to the vehicle's primary planning components which is the main benefit of the presented approach.

The integration of the concept into the \glspl{ADS} of KIT and TUM is ongoing work.
This will allow human-in-the-loop studies in real-world testing to further refine usability and performance of the concept.
An important research question which we will investigate with the presented concept is the impact of a dynamically changing environment on the validity of the remote operator's input and how such changes can be handled safely.
For this, the verification and fallback mechanisms introduced by \textcite{spiekerBetterSafeSorry2024} add safety guarantees for trajectories that have been generated by collaborative planning.

\section*{Acknowledgements}

Nick Le Large and David Brecht, as the first authors, collectively contributed to the content of this work. 
Willi Poh made essential  contributions in the setup and generation of the simulation results.
Jan-Hendrik Pauls, Martin Lauer and Frank Diermeyer made essential contributions to the conception of the research project and revised the paper critically for important intellectual content. 

The authors thank the German Federal Ministry of Education and Research (BMBF)
for being funded in the project \enquote{autotech.agil} (FKZ~01IS22088).

\printbibliography

\end{document}